\documentclass{article}
\usepackage{spconf,amsmath,graphicx}
\usepackage[utf8]{inputenc} 
\usepackage[T1]{fontenc}    
\usepackage{xcolor}
\usepackage{amssymb}

\definecolor{greenForest}{RGB}{34, 139, 34}

\newcommand{\highlightGrey}[1]{%
  \colorbox{gray!50}{$\displaystyle#1$}}

\newcommand{\highlightRed}[1]{%
  \colorbox{red!50}{$\displaystyle#1$}}

\newcommand{\highlightGreen}[1]{%
  \colorbox{greenForest!50}{$\displaystyle#1$}}

\newcommand{\highlightBlue}[1]{%
  \colorbox{blue!50}{$\displaystyle#1$}}

\title{Quaternion convolutional neural networks \\ for heterogeneous image processing}
\name{ Titouan Parcollet$^{1,2}$, Mohamed Morchid$^1$, Georges Linarès$^1$}

\address{$^1$Université d'Avignon, LIA, France\\
  $^2$Orkis, Aix en provence, France\\
 }

\begin{document}
%
\maketitle
\begin{abstract}

Convolutional neural networks (CNN) have recently achieved state-of-the-art results in various applications. In the case of image recognition, an ideal model has to learn independently of the training data, both local dependencies between the three components (R,G,B) of a pixel, and the global relations describing edges or shapes, making it efficient with small or heterogeneous datasets. Quaternion-valued convolutional neural networks (QCNN) solved this problematic by introducing multidimensional algebra to CNN. This paper proposes to explore the fundamental reason of the success of QCNN over CNN, by investigating the impact of the Hamilton product on a color image reconstruction task performed from a gray-scale only training. By learning independently both internal and external relations and with less parameters than real valued convolutional encoder-decoder (CAE), quaternion convolutional encoder-decoders (QCAE) perfectly reconstructed unseen color images while CAE produced worst and gray-scale versions. 

\end{abstract}
%

%
%
\section{Introduction}
\label{sec:intro}

Neural network models are at the core of modern image recognition methods. Among these models, convolutional neural networks \cite{kim2014convolutional}(CNN) have been developed to consider both basic and complex patterns in images, and achieved top of the line results in numerous challenges \cite{he2016deep}. Nonetheless, in the specific case of image recognition, a good model has to efficiently encode local relations within the input features, such as between the Red, Green, and Blue (R,G,B) channels of a single pixel, as well as structural relations, such as those describing edges or shapes composed by groups of pixels. In particular, traditional real-valued CNNs consider pixels as three different and separated values (R, G, B), while a more natural representation is to process a pixel as a single multidimensional entity. More precisely, both internal and global hidden relations are considered at the same level during the training of CNNs.  

Thereby, and strong of many applications \cite{sangwine1996fourier,pei1999color,aspragathos1998comparative}, quaternion neural networks \cite{arena1994neural,arena1997multilayer,isokawa2003quaternion} (QNN) have been proposed to encapsulate multidimensional input features. Quaternions are hyper-complex numbers that contain a real and three separate imaginary components, fitting perfectly to three and four dimensional feature vectors, such as for image processing. Indeed, the three components (R,G,B) of a given pixel are embedded in a quaternion, to create and process pixels as entities. With the purpose to solve the above described problem of local and global dependencies, deep quaternion convolutional neural networks \cite{parcollet2018qcnn,chase2017quat,zhu2018quaternion} (QCNN) have been proposed. In the previous works, better image classification results than real-valued CNN are obtained with smaller neural networks in term of number of parameters. The authors claim that such better performances are due to the specific quaternion algebra, alongside with the natural multidimensional representation of a pixel. Nonetheless, and despite promising results, no clear intuitions of QCNN performances in image recognition have been demonstrated yet. Moreover, these studies employ color images for training and validation sub-processes. 

Therefore, the paper proposes: 1) to explore the impact of the \textit{Hamilton product} (Section \ref{subsec:hamilton}), which is at the heart of the better learning and representation abilities of QNN; 2) to show that quaternion-valued neural networks are able to perfectly learn color features dependencies (R,G,B). Quaternion and real-valued neural networks are therefore compared on a gray-scale to color image task that highlights the capability of a model to learn both internal (i.e. the relations that exist inside a pixel) and external relations of an image. In this extend, a quaternion convolutional encoder-decoder (QCAE) (Section \ref{sec:qcae}) \footnote{Code is available at \url{https://github.com/Orkis-Research/Pytorch-Quaternion-Neural-Networks}} and a real-valued convolutional encoder-decoder \cite{vincent2008extracting} (CAE) are trained to reconstruct a unique gray-scale image from the KODAK PhotoCD dataset (Section \ref{subsec:task}). During the validation process, an unseen color image is presented to both models, and reconstructed pictures are compared visually and with the peak signal to noise ratio (PSNR) as well as the structural similarity (SSIM) metrics (Section \ref{subsec:results}). To validate the learning of internal dependencies, these models must reconstruct the color image without prior information about the color space given from the training phase. The experiments show that QCAE succeeds to produce an almost perfect copy of the testing image, while the CAE fails, by reconstructing a slightly worst and black and white version. Such behavior makes quaternion-valued models a better fit to image recognition in heterogeneous conditions. Indeed, quaternion-valued are less harmed by smaller and  heterogeneous data, due to their ability to dissociate internal and global dependencies trough the \textit{Hamilton product}, and convolutional process respectively. Finally, it is worth noticing that these performances are observed with a reduction of the number of neural parameters of four times for QCAE compared to CAE.     

\begin{figure*}[h!]
 \begin{center}
\scalebox{0.93}{
\includegraphics[width=1\textwidth,origin=c]{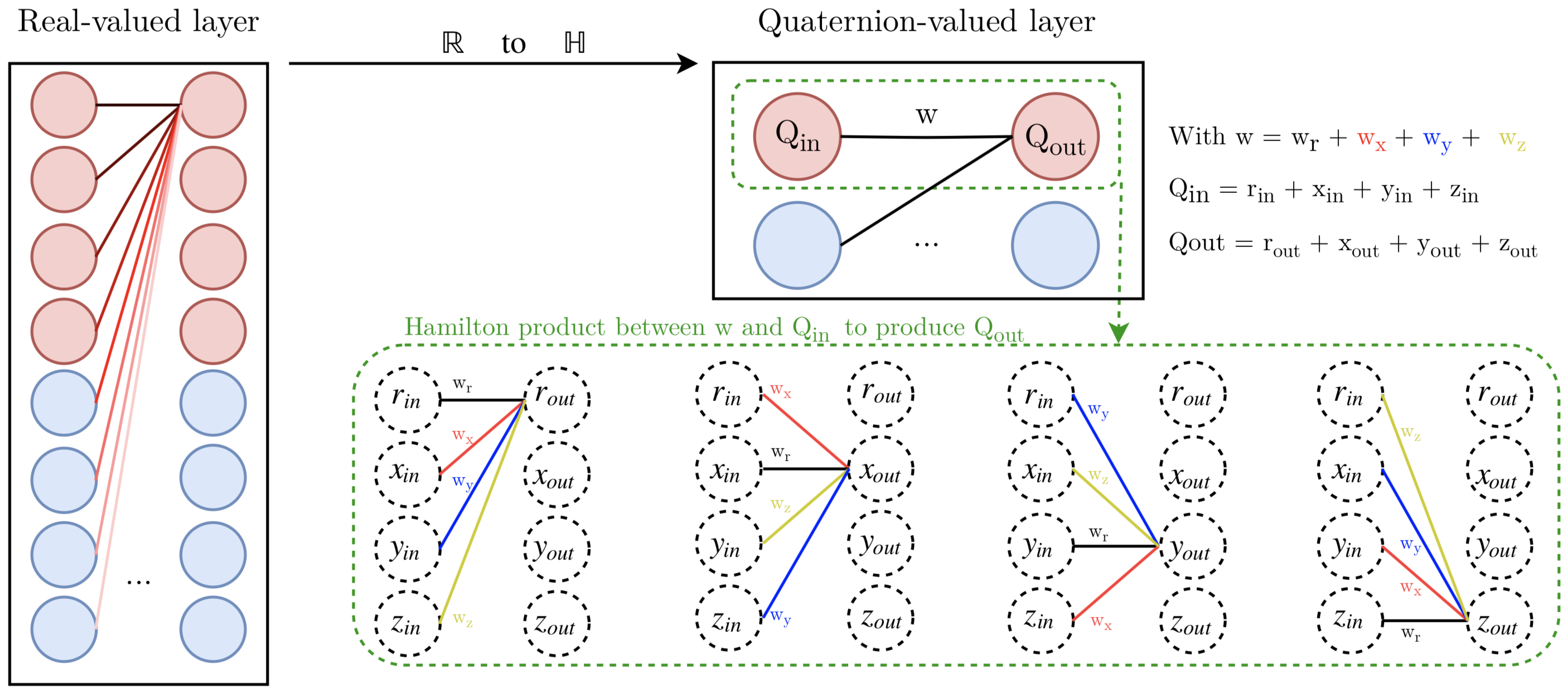}
}
\caption{Illustration of the impact of the \textit{Hamilton product} in a quaternion-valued neural network layer, compared to a traditional real-valued neural network layer}\label{fig:layer}
\end{center}
\end{figure*}

%
%
\section{Quaternion Algebra}
\label{sec:qalgebra}

The quaternion algebra $\mathbb{H}$ defines operations between quaternion numbers. A quaternion Q is an extension of a complex number defined in a four dimensional space as:
\begin{align}
Q = r1 + x\textbf{i} + y\textbf{j} + z\textbf{k},
\end{align}
where $r$, $x$, $y$, and $z$ are real numbers, and $1$, \textbf{i}, \textbf{j}, and \textbf{k} are the quaternion unit basis. In a quaternion, $r$ is the real part, while $x\textbf{i}+y\textbf{j}+z\textbf{k}$ with $\textbf{i}^2=\textbf{j}^2=\textbf{k}^2=\textbf{i}\textbf{j}\textbf{k}=-1$ is the imaginary part, or the vector part. Such a definition can be used to describe spatial rotations.

%
%
\subsection{Hamilton product} 
\label{subsec:hamilton}

The \textit{Hamilton product } ($\otimes$) is used in QNN to remplace the standard real-valued dot product, and to perform transformations between two quaternions $Q_1$ and $Q_2$ following: 
\begin{align}
Q_1 \otimes Q_2=&(r_1r_2-x_1x_2-y_1y_2-z_1z_2)+\nonumber \\
			&(r_1x_2+x_1r_2+y_1z_2-z_1y_2) \boldsymbol i+\nonumber \\
            &(r_1y_2-x_1z_2+y_1r_2+z_1x_2) \boldsymbol j+\nonumber \\
            &(r_1z_2+x_1y_2-y_1x_2+z_1r_2) \boldsymbol k.
\label{eq:hamilton}
\end{align}
The {\em Hamilton product} allows quaternion neural network to capture internal latent relations within the features of a quaternion (see Figure \ref{fig:layer}). In the case of a quaternion-valued neural network, the quaternion-weight components are shared through multiple quaternion-input parts during the {\em Hamilton product }, creating relations within the elements. Indeed, Figure \ref{fig:layer} shows that, in a real-valued neural network, the multiple weights required to code latent relations within a feature are considered at the same level as for learning global relations between different features, while the quaternion weight $w$ codes these internal relations within a unique quaternion $Q_{out}$ during the {\em Hamilton product} (right).

%
%
\section{Quaternion convolutional encoder-decoder}
\label{sec:qcae}

The QCAE is an extension of the well-known real-valued convolutional networks (CNN) \cite{he2016deep} and convolutional encoder-decoder \cite{theis2017lossy} to quaternion numbers. Encoder-decoder models are simple unsupervised structures that aim to reconstruct the input feature at the output \cite{vincent2008extracting}. In a CAE or QCAE, encoding dense layers are simply replaced with convolutional ones, while decoding dense layers are either changed to transposed or upsampled convolutional layers \cite{dumoulin2016guide}. In this extend, let us recall the basics of the quaternion-valued convolution process \cite{chase2017quat,parcollet2018qcnn}. The latter operation is performed with the real-number matrices representation of quaternions. Therefore, a traditional $1D$ convolutional layer, with a kernel that contains $K \times K$ feature maps, is split into 4 parts: the first part equal to $r$, the second one to $x\textbf{i}$, the third one to $y\textbf{j}$ and the last one to $z\textbf{k}$ of a quaternion $Q = r1+x\textbf{i}+y\textbf{j}+z\textbf{k}$. The backpropagation is ensured by differentiable cost and activation functions that have already been investigated for quaternions in \cite{xu2017learning} and \cite{nitta1995quaternary}. As a result, the so-called \textit{split} approach \cite{isokawa2003quaternion,arena1994neural,parcollet2018qcnn,parcollet2016quaternion} is used as a quaternion equivalence of real-valued activation functions. Then, let $\gamma_{ab}^l$ and $S_{ab}^l$ , be the quaternion output and the pre-activation quaternion output at layer $l$ and at the indexes $(a,b)$ of the new feature map, and $w$ the quaternion-valued weight filter map of size $K \times K$. A formal definition of the convolution process is: 
\begin{align}
\gamma_{ab}^l &=\alpha(S_{ab}^l),
\end{align}
with
\begin{align}
\label{eq:forward}
S_{ab}^l&=\sum\limits_{c=0}^{K-1}\sum\limits_{d=0}^{K-1}w^l \otimes \gamma_{(a+c)(b+d)}^{l-1},
\end{align}
where $\alpha$ is a {\em quaternion split activation} function \cite{isokawa2003quaternion,arena1994neural,parcollet2018qcnn,parcollet2016quaternion}. The output layer of a quaternion neural network is commonly either quaternion-valued such as for quaternion approximation \cite{arena1997multilayer}, or real-valued to obtains a posterior distribution based on a softmax function following the split approach. Indeed, target classes are often expressed as real numbers.

%
%
\section{Experiments and results}
\label{sec:pagestyle}

This section details the experiments (Section \ref{subsec:task}), the models architectures (Section \ref{subsec:models}), and the results (Section \ref{subsec:results}) obtained with both QCAE and CAE on a gray to color task with the KODAK PhotoCD dataset.

%
%
\subsection{From gray-scale to the color space}
\label{subsec:task}

We propose first to highlight the ability of a model to learn the internal relations that compose pixels ({\it i.e.} the color space), and ensures the robustness of the model in heterogeneous training/validation conditions. In this extend, models are trained to compress and reproduce a unique gray-scale image in an encoder-decoder fashion, and are then fed with two different color images at validation time. Models are expected to reproduce the exact same colors than the original test samples. Experiments are based on the KODAK PhotoCD data-set~\footnote{http://r0k.us/graphics/kodak}. A random image (See Figure \ref{fig:results}) is converted to gray-scale following the basic luma formula \cite{hamilton2004jpeg} and used as the training sample, while the others original color images are used as a validation subset. Therefore, training is performed on a single gray-scale image of $512\times768$ pixels with the gray value of a given pixel $p_{x,y}$ repeated three times to compose a quaternion $Q(p_{x,y})=0+\highlightGrey{GS(p_{x,y})}\textbf{i}+\highlightGrey{GS(p_{x,y})}\textbf{j}+\highlightGrey{GS(p_{x,y})}\textbf{k}$. For a fair comparison, the gray value is also concatenated three times for each pixel in the real-valued CNN. Finally, the quaternion $Q(p_{x,y})=0+\highlightRed{R(p_{x,y})}\textbf{i}+\highlightGreen{G(p_{x,y})}\textbf{j}+\highlightBlue{B(p_{x,y})}\textbf{k}$ based on color images is composed and processed at validation time, while $R,G,B$ components are concatenated for CNN. Reconstructed pictures are evaluated visually and with the peak signal to noise ratio (PSNR) \cite{turaga2004no} as well as structural similarity (SSIM)\cite{wang2004image} metrics.

%
%
\subsection{Models architectures}
\label{subsec:models}

QCAE and CAE have the same topology. It is worth noticing that the number of output feature maps is four times larger in the QCAE due to the quaternion convolution, meaning $8$ quaternion-valued feature maps correspond to $32$ real-valued ones. Therefore, each model has two convolutional encoding layers and transposed convolutional decoding layers that deal with the same dimensions, but with different internal sizes. Indeed quaternion features maps are of size $8$ and $16$ to deal with an equivalently size of $32$ and $64$ for the CAE. Such dimensions ensure an encoding dimension slightly smaller than the original picture size. Kernel size and strides are set to $3$ and $2$ across all the layers respectively. Training is performed during $3,000$ epochs with the Adam optimizer \cite{kingma2014adam}, vanilla hyper-parameters and a learning rate of $5e^{-4}$. The hardtanh \cite{collobert2004large} activation function is used in both convolutional and transposed convolutional layers. Finally, quaternion parameters are initialized following the proposal of \cite{parcollet2018qrnn}.

%
%
\subsection{Results and discussions}
\label{subsec:results}

\begin{figure*}[h!]
 \begin{center}
\scalebox{0.98}{
\includegraphics[width=1\textwidth,origin=c]{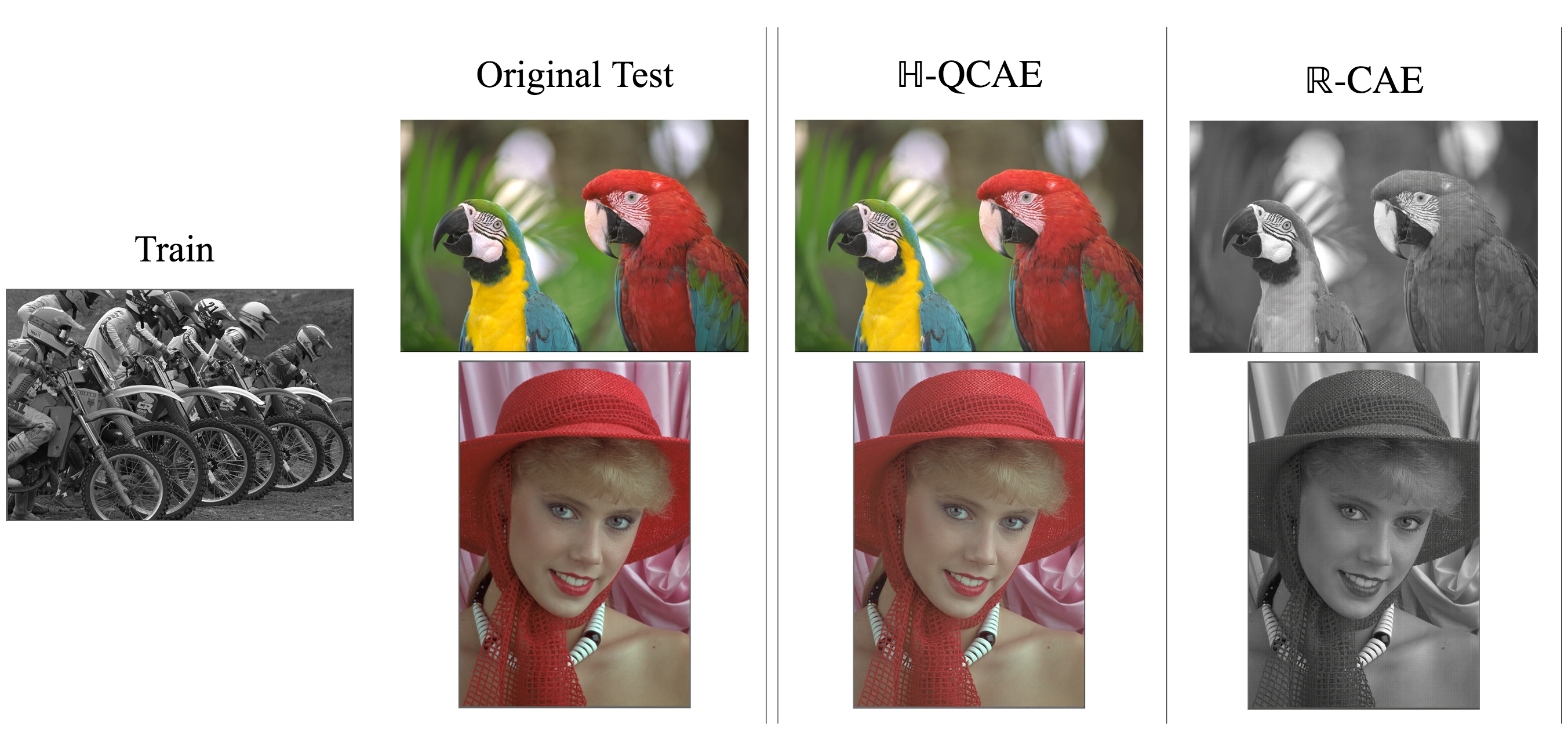}
}
\caption{Results on the gray-scale to color task with the KODAK data-set. A gray-scale training picture (\textit{Train}) and two coloured original test images (\textit{Original Test}) are randomly selected on the KODAK data-set and reproduced by both QCAE and CAE.}\label{fig:results}
\end{center}
\end{figure*}

The results are reported in Figure \ref{fig:results}. It is first important to notice that quaternion-valued CAE produced almost perfect color images {\it w.r.t.} to the test, while CAE completely failed to capture colors by outputting a black and white version. As motivated in Section \ref{sec:qalgebra}, the quaternion representation alongside with the \textit{Hamilton product} force the QCAE to consider and preserve the internal latent relations between the components of a quaternion ({\it i.e.} a pixel). Consequently, QCAE easily captures the color space from a gray-scale image since it learns to produce the exact same values from the input at the output, while real-valued CAE learns a gray-scale mapping by generating three identical components. 

Other numerical measures are obtained based on the PSNR and SSIM of the reconstructed pictures. Due to the fact that CAE fails to learn colors, we propose to compare CAE results to the gray-scale equivalent of the test pictures. QCAE results are compared to the true color images. Consequently, we can measure how good each model is to reconstruct testing images, without being biased by the fact that CAE fails to learn colors. QCAE obtains a PSNR of $31.68$dB and $28.06$dB compared to $29.95$dB and $27.01$dB obtained with the CAE for the parrots and women images respectively. Nonetheless, while PSNR measures the amount of noise contained in an image, SSIM reports the structural and visual correlation of two pictures. SSIM of $0.96$ and $0.93$ are reported for QCAE compared to $0.87$ and $0.86$ for the CAE on the parrots and women images respectively. QCAE offered a better reconstructed image quality in both PSNR and SSIM metrics, even considering the inability of CAE to learn the color space. 
Moreover, the QCAE is composed of $6.4$K parameters compared to $25$K for the CAE. It is easily explained by the quaternion algebra. In the case of a dense layer with $1,024$ input values and $1,024$ hidden units, a real-valued model will have $1,024^2\approx1$M parameters, while to maintain equal input and output nodes ($1,024$) the quaternion equivalent has $256$ quaternions inputs and $256$ quaternion-valued hidden units. Therefore the number of parameters for the quaternion model is $256^2\times4\approx0.26$M. 

\noindent
\textbf{Discussions.} In the one hand, the size reduction offered by QNN turns out to produce better results with an higher generalization capacity and may have other advantages such as a smallest memory footprint while saving models. Then, the natural internal relation representation induced by the \textit{Hamilton product}, alongside with the convolution process provide an important step toward better performances of models that operate in heterogeneous contexts, or with very small data-sets. The small number of neurons allows the QCAE to obtain ``robust'' and ``compact'' memory that build a robust hidden representations of the image content in the latent space. Indeed, QCAE are not altered by heterogeneous color spaces ({\it e.g.} corpus of boats with a predominating blue spectrum), and are able to learn internal relations with very few examples trough the \textit{Hamilton product}.

%
%
\section{Conclusion}
\label{sec:conclusion}

This paper proposes to clarify the recent better performances observed on image recognition with quaternion-valued neural networks, through a investigation of the impact of the \textit{Hamilton product}. The conduced experiments demonstrate that quaternion convolutional encoder-decoders are able to perfectly learn the color-space with a training performed on a unique gray-scale image, while real-valued CAE fail, proving that the \textit{Hamilton product} allows QNN to encode local dependencies such as the RGB relation of a pixel. Moreover, QCAE produce better quality reconstructions with respect to the PSNR and SSIM metrics than CAE, even with considering the unability of CAE to learn colors. Moreover, the quaternion representation offers more compact and expressive models. Thereby, the experiments have validated the initial intuition that the \textit{Hamilton product}, alongside with the convolution process, allow QCAE to better separate both local and global dependencies of color images. These results are an important step forward for a more robust image recognition system in heterogeneous conditions. Future work will attempts to introduce the efficient quaternion representation to image compression and image recognition.  

\bibliographystyle{IEEEbib}

\end{document}